# Admissible Time Series Motif Discovery with Missing Data

Yan Zhu, ‡Abdullah Mueen, Eamonn Keogh
University of California, Riverside, ‡University of New Mexico
yzhu015@ucr.edu, eamonn@cs.ucr.edu, ‡mueen@cs.unm.edu

**Abstract**

The discovery of time series motifs has emerged as one of the most useful primitives in time series data mining. Researchers have shown its utility for exploratory data mining, summarization, visualization, segmentation, classification, clustering, and rule discovery. Although there has been more than a decade of extensive research, there is still no technique to allow the discovery of time series motifs in the presence of missing data, despite the well-documented ubiquity of missing data in scientific, industrial, and medical datasets. In this work, we introduce a technique for motif discovery in the presence of missing data. We formally prove that our method is *admissible*, producing no false negatives. We also show that our method can "piggy-back" off the fastest known motif discovery method with a small constant factor time/space overhead. We will demonstrate our approach on diverse datasets with varying amounts of missing data.

## 1  INTRODUCTION

Time series motifs are short approximately repeated patterns within a longer time series dataset. The fact that such patterns are *conserved* often suggests underlying structure and regularities that can be exploited in many ways. Some examples include: rule discovery [16][13], forecasting [10], or building better classifiers [2]. However, despite over a decade of active research, there is no known method to allow the discovery of the motif in the presence of missing data.

Paradoxically, in spite of improvements in sensor technology, missing data is becoming *more* prevalent. This is because sensors are now so cheap that we are willing to deploy them in hostile environments with intermittent and unreliable bandwidth [17]. In many cases, sensors have become "throwable" and disposable [7].

**Figure 1** shows an example of a motif in a music processing domain [12]. Note that both occurrences of the motif contain sections of missing data.

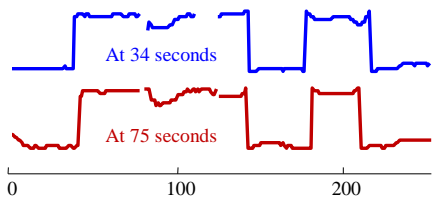

**Figure 1:** A four-second long motif that appears in the pitch contour time series of a Cypriot folk song, *Kotsini Trantafillia* (Red Rose-tree). Note that both occurrences have multiple instances of missing data [12].

Here the data could be missing for one of two reasons. It *might* be meaningful, i.e., the vocalist may not have produced a sound at these times. It is also possible that the missing data reflects poor audio quality, loose wires, etc. We are agnostic to such issues in this work. We simply note that in either case, there is no motif discovery algorithm defined in the literature for such data.

More generally, in time series data mining, missing data is often handled by filling in the missing values with some interpolation or imputation method [8] and running the analysis unaltered. However, as we shall demonstrate, no matter what interpolation or imputation method is used, and no matter what current motif discovery algorithm is used, this can result in false negatives.

To solve this problem, we introduce a novel algorithm that does not allow false negatives. The method *may* allow occasional false positives, but since the discovered motifs are typically examined by the human eye [11][13][25], or some subsequent analysis, the false positives (if any) can be filtered out at a later stage.

We call our algorithm Motif Discovery with Missing Data, MDMS. Our MDMS algorithm is built on top of the recently introduced Matrix Profile data structure [22]. Matrix Profile-based motif discovery has been shown to have following advantageous features, all of which are inherited by our proposed MDMS:

- It is simple and parameter-free. In contrast, the more general motif discovery algorithms require building and tuning spatial access methods and/or hash functions.
- It requires an inconsequential space overhead, just O($n$) with a small constant factor.
- While the exact algorithm is scalable, for extremely large datasets we can compute the results in a timely fashion, allowing ultra-fast *approximate* solutions.
- Having computed the motifs for a dataset, we can incrementally update them efficiently. Thus for reasonable arrival rates (< 20Hz), we can effectively maintain exact motifs on streaming data forever.
- The algorithm is easy to parallelize, both on multicore processors and in distributed systems.

The rest of the paper is organized as follows. Section 2 reviews related work and introduces the necessary background materials and definitions. In Section 3, we introduce our algorithms. Section 4 gives a detailed empirical evaluation. Finally, in Section 5 we offer conclusions and directions for future work.

## 2  RELATED WORK AND BACKGROUND

Time series motif discovery has been an active area of research for over a decade [3][11][25]. However, to the best of our knowledge, there is no existing algorithm that can find motifs in the presence of missing data. In principle, *any* motif discovery algorithm could be used with missing data, if we use some imputation algorithm to fill in the missing values. There are hundreds of imputation algorithms in the literature (see [21] and the references therein) to choose from, but no matter which one we use, we may obtain false negatives with respect to the *oracle* data (the true underlying data, without missing values).

To see this let us consider a simple example. Suppose we have a dataset that is composed of just three (sub)sequences:

A={0,2,0,2}, B={0,2,0,2}, C={0,-1,0,2}

Clearly the pair A|B is a perfect motif. Now suppose that the second value in A is missing. The most obvious imputation technique is interpolation from the two neighbors of the missing data point, giving us $A_{miss}$ = {0,0,0,2}. As we can see in **Figure 2**, this one change means that we no longer discover the pair A|B as a perfect

motif, but instead we are lead to believe that the pair A|C is the best motif in the dataset.

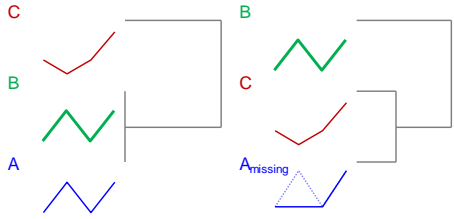

**Figure 2:** *left*) A contrived dataset in which the pair A|B is a perfect motif. *right*) If A had its second value replaced by the most common imputation algorithm, we would fail to discover A|B as the motif.

While this is a trivial example, it is easy to see that no matter what imputation algorithm is used, using a simple adversarial augment we can always construct datasets for which the classic motif algorithms would produce false negatives. This problem is even more severe with complex datasets that contain a lot of high-frequency patterns and noise, for example seismology datasets.

In [26] the authors showed two repeating earthquake patterns that appeared approximately 14 years apart. We have found that if there were just a handful of missing data points in one of the earthquake samples, we would be unable to detect a match between them with any common imputation method, as such high-frequency and noisy data defies the assumptions that most imputation techniques assume.

**Figure 2** showed that imputation methods can produce possible false negatives even if we have a random single value missing in the data. Moreover, another disadvantage of current imputation methods is that they cannot predict *block-missing data* well [24]. In some circumstances, due to malfunction of the sensor or other anomaly factors, we may lose reading from a sensor at consecutive timestamps (instead of sparse, single missing timestamps at random locations of the data). This is often called *block-missing* data. In [24] the authors proposed a state-of-the-art spatial-temporal imputation method to predict the block-missing data by learning from not only the real-valued reading of the same sensor, but also from the reading of several geographically nearby sensors. The information from other sensors greatly improved their imputation accuracy. However, this method does not apply when we only have access to one single sensor, or when all the sensors contain block-missing data at the same timestamps due to regional power outage or communication errors.

Finally, note that our problem is not artificial or contrived in anyway. The literature is replete with examples of data analysts frustrated by the inability to perform motif discovery in the presence of missing data. For example, a recent paper studies recurrent water consumption behavior by Australian consumers [19]. The authors observe "*A small proportion of all hourly readings are missing…, probably due to server failure or maintenance*". The authors observe "*A small proportion of all hourly readings are missing…, probably due to server failure or maintenance.*" The authors realize that any imputation method used here has a risk of producing false negatives by noting that "*…hourly water consumption is highly unpredictable, we ignored the points of missing hours for the routine discovery, rather than approximating missing readings*." However, their solution of ignoring some data runs the risk of missing interesting patterns.

### 2.1 Dismissing Apparent Solutions
In this section, we continue the discussion of related work, while explicitly dismissing some proposed solutions to the task at hand.

The last decade has seen several distance measures for handling uncertain time series, including PROUD [23], DUST [15], PBRQ, MUNICH etc. One might believe that these measures could be used to replace the Euclidean Distance subroutine in an existing motif discovery algorithm. However, we believe this is not possible for the following reasons:

- These methods assume not *missing* data, but *uncertain* data. For example, they can address the situation where an observation is not known precisely but comes from some known distribution. In [15] for example, they explicitly model the normal, exponential, and uniform error distributions. However, for generality, we do not wish to make such strong assumptions.

- Even if we assume that we could somehow avail of an existing uncertain distance measure, none of them are metrics (only *measures*). However, all speedup techniques for exact motif discovery that we are aware of require and exploit metric properties [3][11]. This suggests we must resort to a $O(n^2m)$ brute force search ($n$ is the length of the time series, $m$ the motif length). However, our proposed method is $O(n^2)$. As $m$ may be in the thousands (see Section 4), this suggests a three order of magnitude time difference.

- Finally, we want to be able to guarantee that our search produces no false negatives in the face of missing data. To the best of our knowledge, no existing uncertain time series similarity measures can support this requirement.

The reader may not appreciate why our task-at-hand is hard, because the analogue problem with *strings* is trivial. Suppose we are asked to compare the following text strings "*Norwegian blue*" and "*Norwegian wood*" under the Hamming Distance, and we encounter missing values, represented here by "*".

```
Norwe*ian_blu*
N*rwe*ian_wood
```

We can easily compute both the lower and upper bound of the Hamming Distance. In the former case, we would assume that all the missing values in one word are the *same* as their counterparts in the other word. Given that the only letters we can be sure are differing are "woo" vs "blu," and we have a lower bound of three. In the latter case, we would assume that all the missing values in one word are *different* than their counterparts in the other word. These three pessimistic differences combined with the three observed differences give us an upper bound of six.

However, consider the time series version of this problem. Suppose we have the following two time series:

```
[ 0.5, 0.1, ***, ***, -0.6, -0.7, 0.0 ]
[ 0.3, 0.1, 1.1, ***, -0.6,  ***, 0.1 ]
```

One might consider applying similar logic here. For example, accumulating 0.2 error (i.e. |0.5 - 0.3|) from the first pair of numbers, then 0.0 error (i.e. |0.1 - 0.1|) from the second pair, etc. However, the critical difference is that the time series must be *normalized* before comparison. This is because, aside from the rare and well understood exceptions [5], it is meaningless to compare time series without normalizing them first. This presents a problem as z-normalization (the most common normalization technique [4]) requires us to know the exact mean and standard deviation of the data, which are undefined when we have a single missing data point.

Thus, for any pair of corresponding *known* points (for example the 0.5 and the 0.3 in the above), it is possible that the true (had we known the mean and standard deviation of the data to allow the correct normalization) difference between them could stay the same, increase, or decrease by arbitrary amounts. This suggests that producing tight upper or lower bounds will be nontrivial.

## 2.2 Definitions and Notations

In this section, we introduce all of the necessary notations and definitions needed to explain our solution. Some of the text is redundant with [26], but is included for completeness.

We begin by defining the data type of interest, *time series*:

**Definition 1:** A *time series* $T$ is a sequence of real-valued numbers $t_i$: $T = t_1, t_2, ..., t_n$, where $n$ is the length of $T$.

For motif discovery, we are not interested in the *global* properties of time series, but in the *local* regions, known as *subsequences*:

**Definition 2:** A *subsequence* $T_{i,m}$ of a time series $T$ is a continuous subset of the values from $T$ of length $m$ starting from position $i$. Formally, $T_{i,m} = t_i, t_{i+1},..., t_{i+m-1}$, where $1 \leq i \leq n-m+1$.

We can generalize both definitions to allow for the possibly that at least one value is missing. For clarity, we differentiate such time series (including subsequences) with a "bar." To keep both in a common notation and for implementation purposes (i.e. Matlab, etc.), we use *NaN*s as a placeholder for missing values.

**Definition 3:** A *missing value time series* $\overline{T}$ is a sequence of values that are either real-valued numbers or *NaN*s, $\overline{t}_i$: $\overline{T} = \overline{t}_1, \overline{t}_2, ..., \overline{t}_n$, where $n$ is the length of $\overline{T}$.

In the rest of the paper, we assume that $T$ is the *actual* time series of $\overline{T}$ before the missing values were created by some process. That is to say, we would have obtained $T$ instead of $\overline{T}$ if the sensors were functioning properly. We do not know $T$ precisely, but we have $\overline{t}_i = t_i$ ($1 \leq i \leq n$) if $\overline{t}_i \neq NaN$.

The following restates definitions introduced with the STOMP framework [26], but are reviewed for clarity.

We can take any subsequence from a time series and compute its distance to *all* sequences. We call an ordered vector of such distances a *distance profile*:

**Definition 4:** A *distance profile* $D_i$ of time series $T$ is a vector of the Euclidean distances between a given query subsequence $T_{i,m}$ and each subsequence in the time series $T$. Formally, $D_i = [d_{i,1}, d_{i,2},..., d_{i,n-m+1}]$, where $d_{i,j}$ ($1 \leq i, j \leq n-m+1$) is the distance between $T_{i,m}$ and $T_{j,m}$.

We assume that the distance is measured by Euclidean distance between z-normalized subsequences [4]. Unless otherwise stated, in the rest of the paper, we always use $d_{i,j}$ to represent the z-normalized Euclidean distance between $T_{i,m}$ and $T_{j,m}$.

We are interested in finding the nearest neighbors of all subsequences in $T$, as the closest pairs are the classic definition of time series motifs [3][11]. Note that by definition, the $i^{th}$ location of distance profile $D_i$ is zero and very close to zero just before and after this location. In the literature, such matches are called *trivial matches* [11]. We avoid such matches by ignoring an "exclusion zone" of length $m/4$ before and after the location of the query. In practice, we set $d_{i,j}$ to infinity ($i-m/4 \leq j \leq i+m/4$) while evaluating $D_i$.

We use a vector called a *matrix profile* to represent the distances between all subsequences and their nearest neighbors:

**Definition 5:** A *matrix profile* $P$ of time series $T$ is a vector of the Euclidean distances between each subsequence $T_{i,m}$ and its nearest neighbor (closest match) in time series $T$. Formally, $P = [\min(D_1), \min(D_2),..., \min(D_{n-m+1})]$, where $D_i$ ($1 \leq i \leq n-m+1$) is the distance profile $D_i$ of time series $T$.

This vector is called the matrix profile, because one way to compute it would be to compute the full distance matrix of all pairs of subsequences in time series $T$, and then evaluate the minimum value of each column. **Figure 3** illustrates both a *distance profile* and a *matrix profile* created on the same raw time series $T$.

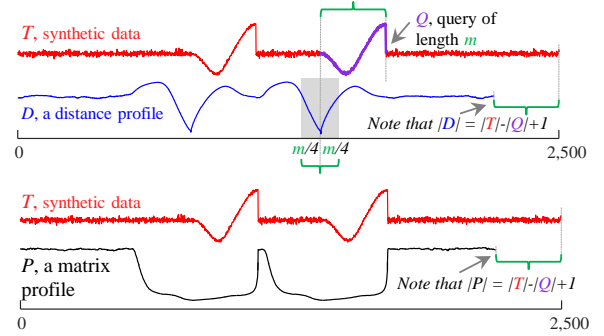

**Figure 3:** *top*) One distance profile (Definition 4) created from a random subsequence $Q$ of $T$. If we created distance profiles for all possible subsequences of $T$, the element-wise minimum of this set would be the matrix profile (Definition 5) shown at (*bottom*). Note that the two lowest values in $P$ are at the location of the 1st motif [3][11].

The $i^{th}$ element in the matrix profile $P$ tells us the Euclidean distance from the subsequence $T_{i,m}$ to its nearest neighbor in the time series $T$. However, it does not tell us *where* that neighbor is located. This information is recorded in a companion data structure called the *matrix profile index*.

**Definition 6:** A *matrix profile index* $I$ of time series $T$ is a vector of integers: $I=[I_1, I_2, ... I_{n-m+1}]$, where $I_i=j$ if $d_{i,j} = \min(D_i)$.

By storing the neighboring information this way, we can efficiently retrieve the nearest neighbor of query $T_{i,m}$ by accessing the $i^{th}$ element in the matrix profile index.

For clarity of presentation, we have confined this work to the single dimensional case; however, nothing about our work intrinsically precludes generalizations to multidimensional data.

To briefly summarize this section: we can create two meta time series, the *matrix profile* and the *matrix profile index*, to annotate a time series $T$ with the distance and location of all its subsequences' nearest neighbors within itself. For a time series $T$ of length $n$, the recently introduced STOMP algorithm [26] is able to compute the two meta files with a mere $O(n^2)$ time complexity and $O(n)$ space complexity, which enables a fast motif discovery in a massive time series. However, the STOMP algorithm is not applicable to any time series $\overline{T}$ with missing values (*NaN*s). Here we claim our MDMS algorithm can solve the missing data problem with the same time and space complexity. We leave the detailed discussion of the algorithm to Section 3.

## 2.3 Pseudo Missing Data

Before introducing our solution to the missing data problem, we take the time to point out that the problem is more general than one may assume when we consider the generalization of the Pseudo Missing Data (PMD). Informally, we define PMD as data that technically is not missing, but *effectively* is. **Figure 4** illustrates three kinds of PMD frequently encountered.

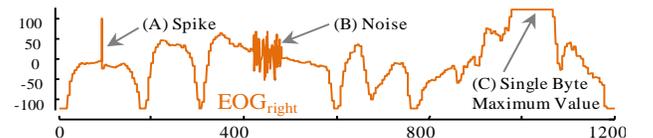

**Figure 4:** A snippet of an Electrooculogram (EOG) exhibits three kinds of pseudo missing data

In **Figure 4**.A, we see a "spike." Given what we know about this domain, it is inconceivable that the human eye could move fast enough to produce this data, so it is clearly an artifact. Likewise, in **Figure 4**.B, the dramatic increase in variance suggests that this section of data is not likely to faithfully represent the underlying physical process. Finally, in **Figure 4**.C the perfectly flat plateau is not reflective of reality, but is simply a region where the physical process exceeds the 8-bit precision available to record it.

In all three cases, the best thing to do would be to treat the data as missing. Note that this decision is domain-dependent; there are clearly domains where a short spike represents some physical event, or where a perfectly flat plateau represents a physical limitation, which is not a quirk of the hardware/software use.

## 3 ALGORITHMS
### 3.1 An Intuitive Preview

We begin by previewing our solution. As shown in [22][26], if $T$ is used to compute a matrix profile then finding the motifs is trivial. The location of the *smallest* pair of values in the matrix profile is also the location in $T$ of the optimal motif pair. Moreover, other definitions of motifs, such as the top-K motifs or range motifs [11][22], can also be easily extracted from the matrix profile. Given this, our solution to the missing data problem is to build a special matrix profile using $\overline{T}$. This special matrix profile will be very similar to the true matrix profile, and be a (in general, *very* tight) lower bound for it. If we use the existing motif extraction algorithms [22][26] to pull out motifs from this matrix profile, we may have false positives, but we will have no false negatives [5]. Thus most of our contribution outlined below is to show how we can build this special matrix profile.

### 3.2 Lower Bound Matrix Profile

To create the special matrix profile data structure, our MDMS algorithm evaluates the z-normalized Euclidean distance between every pair of subsequences within a missing value time series $\overline{T}$. Depending on whether or not the subsequences have missing values, we may encounter three different situations. Assume that the pair of subsequences under consideration is $\overline{T}_{i,m}$ and $\overline{T}_{j,m}$.

- **Case 1**: Neither $\overline{T}_{i,m}$ nor $\overline{T}_{j,m}$ contain any missing value (*NaN*). Normally, this applies to most subsequence pairs within time series $\overline{T}$ if $\overline{T}$ contains more real-valued numbers than *NaN*s. The traditional exact z-normalized Euclidean distance between $\overline{T}_{i,m}$ or $\overline{T}_{j,m}$ can be evaluated in this case.
- **Case 2**: $\overline{T}_{i,m}$ contains missing values (*NaN*s) while $\overline{T}_{j,m}$ does not, or vice versa.
- **Case 3**: Both $\overline{T}_{i,m}$ and $\overline{T}_{j,m}$ have missing values (*NaN*s).

In cases 2 and 3, the exact z-normalized Euclidean distance $d_{i,j}$ between $\overline{T}_{i,m}$ or $\overline{T}_{j,m}$ cannot be evaluated. However, we can evaluate a lower bound of the distance between $\overline{T}_{i,m}$ and $\overline{T}_{j,m}$, $d_{i,j}^{LB}$, such that $d_{i,j}^{LB} \leq \min(d_{i,j})$. That is to say, no matter what the missing values in $\overline{T}_{i,m}$ or $\overline{T}_{j,m}$ are, we guarantee that $d_{i,j}$, the actual distance between $\overline{T}_{i,m}$ and $\overline{T}_{j,m}$, is no less than $d_{i,j}^{LB}$.

Now we are ready to redefine **Definition 4** in the context of missing values. We keep those $d_{i,j}$ values corresponding to Case 1 unchanged, and use $d_{i,j}^{LB}$ to replace the $d_{i,j}$ values corresponding to Cases 2 and 3. We will then obtain a *lower bound distance profile*:

**Definition 7:** A *lower bound distance profile* $D_i^{LB}$ of a missing value time series $\overline{T}$ is a vector $D_i^{LB} =[\ \bar{d}_{i,1}\ ,\ \bar{d}_{i,2}\ ,...,\ \bar{d}_{i,n-m+1}\ ]$, where $\bar{d}_{i,j}=d_{i,j}$ ($1 \leq i, j \leq n\text{-}m+1$) if neither $\overline{T}_{i,m}$ nor $\overline{T}_{j,m}$ contains *NaN*s (Case 1), and $\bar{d}_{i,j}=d_{i,j}^{LB}$ ($1 \leq i, j \leq n\text{-}m+1$) otherwise.

Similarly, we can redefine **Definition 5** and **Definition 6** in the context of missing values.

**Definition 8:** A *lower bound matrix profile* $P^{LB}$ of a missing value time series $\overline{T}$ is a vector of the *lower bound* Euclidean distances between each subsequence $\overline{T}_{i,m}$ and its nearest possible neighbor (closest possible match) in $\overline{T}$. Formally, $P^{LB} = [\min(D_1^{LB}), \min(D_2^{LB}),..., \min(D_{n-m+1}^{LB})]$, where $D_i^{LB}$ ($1 \leq i \leq n\text{-}m+1$) is the lower bound distance profile $D_i^{LB}$ of the time series $T$.

**Definition 9:** A *lower bound matrix profile index* $I^{LB}$ of $\overline{T}$ is a vector of integers: $I^{LB} =[I_1^{LB},\ I_2^{LB},\ ...\ I_{n-m+1}^{LB}]$, where $I_i^{LB} =j$ if $\bar{d}_{i,j}= \min(D_i^{LB})$.

The lower bound matrix profile gives us an optimistic approximation of the distance between every subsequence and its nearest possible neighbor in $\overline{T}$. This approximation may produce false positives, but it will not allow false negatives.

We believe this is the best approach for the task at hand: as [26] shows, the cost to filter out false positive motifs is very low once we have the matrix profile, but we cannot afford the occurrence of any false negatives, since they may include the most important patterns in the time series. In the following sections, we will introduce the lower bounds corresponding to Cases 2 and 3 respectively; then we will introduce our MDMS algorithm, which evaluates a lower bound of the matrix profile.

### 3.3 The Lower Bound Euclidean Distance
#### 3.3.1 Case 2

Let us first consider Case 2, where $\overline{T}_{i,m}$ contains missing values (*NaN*s) while $\overline{T}_{j,m}$ does not, or vice versa. Without loss of generality, for now we assume $\overline{T}_{i,m}$ is the subsequence that contains missing values, and $\overline{T}_{j,m}$ is the subsequence without missing values. **Figure 5** shows a visual example of this case.

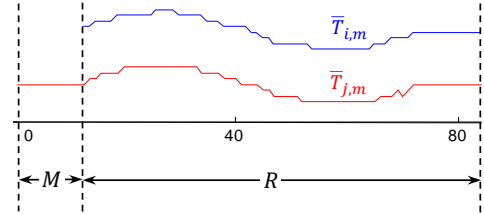

**Figure 5:** *top*) A subsequence with missing values. *bottom*) A subsequence without missing values.

Here $M = \{k\ |\ 1 \leq k \leq m$ and $t_{i+k-1} = NaN\ \}$ is a set of the locations of missing values within $\overline{T}_{i,m}$, and $R = \{k\ |\ 1 \leq k \leq m$ and $t_{i+k-1} \neq NaN$ and $t_{j+k-1} \neq NaN\ \}$ is the intersection of the real-valued locations within $\overline{T}_{i,m}$ and those within $\overline{T}_{j,m}$. To evaluate the lower bound distance of the two subsequences, first we need to z-normalize them [4][11][22][26].

We assume that $\mu_i$ and $\sigma_i$ are the mean and standard deviation of $\overline{T}_{i,m}$, $\mu_j$ and $\sigma_j$ are the mean and standard deviation of $\overline{T}_{j,m}$. Note that because $\overline{T}_{i,m}$ has missing values, we cannot evaluate $\mu_i$ and $\sigma_i$; however, we can treat them as variables. Assume $d_{i,j}$ is the distance between $\overline{T}_{j,m}$ and the oracle subsequence of $\overline{T}_{i,m}$, we can easily obtain a lower bound distance of $d_{i,j}$ by ignoring the missing part $M$ in **Figure 5**:

$$d_{i,j}^{LB} = \sqrt{\min_{\mu_i,\sigma_i} \sum_{k\in R}\left(\frac{t_{i+k-1}-\mu_i}{\sigma_i}-\frac{t_{j+k-1}-\mu_j}{\sigma_j}\right)^2} \quad (1)$$

Assume $f_1 = {d_{i,j}^{LB}}^2$. We can linearly transform $f_1$ as:

$$f_1 = d_{i,j}^{LB^2} = \left(\frac{\sigma_j^R}{\sigma_j}\right)^2 \min_{\mu,\sigma} \sum_{k \in R} \left(\frac{t_{i+k-1}-\mu}{\sigma} - \frac{t_{j+k-1}-\mu_j^R}{\sigma_j^R}\right)^2 \quad (2)$$

$R$ is shown in **Figure 5**. We assume $T_{i,m}^R$ is the real-valued part of $\overline{T}_{i,m}$, $T_{j,m}^R$ is the subset of $\overline{T}_{j,m}$ corresponding to $R$, $\mu_i^R$ and $\sigma_i^R$ are the mean and standard deviation of $T_{i,m}^R$, $\mu_j^R$ and $\sigma_j^R$ are the mean and standard deviation of $T_{j,m}^R$.

**Figure 6** visualizes (2). Note that $T_{j,m}^R$ is z-normalized in (2), so its offset and scale are fixed in **Figure 6**; to obtain $d_{i,j}^{LB}$, we would like to adjust $\mu$ (corresponding to the offset of $T_{i,m}^R$) and $\sigma$ (corresponding to the scale of $T_{i,m}^R$), such that the Euclidean distance between $T_{i,m}^R$ and $T_{j,m}^R$ is minimized.

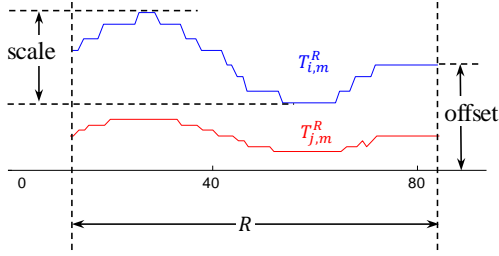

**Figure 6: Different setting of $\mu$ and $\sigma$ changes the offset and the scale of $T_{i,m}^R$. Note that the offset and scale of $T_{j,m}^R$ are fixed.**

By solving $\frac{\partial f_1}{\partial \mu} = 0$ and $\frac{\partial f_1}{\partial \sigma} = 0$, and substituting $\sigma$ and $\mu$ back into (2), we have:

$$d_{i,j}^{LB} = \begin{cases} \frac{\sigma_j^R}{\sigma_j}\sqrt{|R|} & if\ q_{i,j} \leq 0 \\ \frac{\sigma_j^R}{\sigma_j}\sqrt{|R|(1-q_{i,j}^2)} & if\ q_{i,j} > 0 \end{cases} \quad (3)$$

Here $R$ is the intersection of the real-valued locations within $\overline{T}_{i,m}$ and those within $\overline{T}_{j,m}$, $q_{i,j}$ is the Pearson Correlation Coefficient between $T_{i,m}^R$ and $T_{j,m}^R$:

$$q_{i,j} = \frac{\sum_{k \in R} t_{j+k-1} t_{i+k-1} - |R|\mu_i^R \mu_j^R}{|R|\sigma_i^R \sigma_j^R} \quad (4)$$

The analysis of Case 2 is now complete. Let us turn to Case 3, where both $\overline{T}_{i,m}$ and $\overline{T}_{j,m}$ contain missing values.

### 3.3.2 Case 3

**Figure 7** shows an example of Case 3.

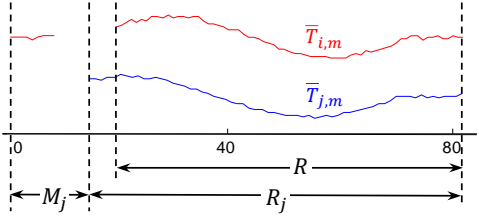

**Figure 7: Two subsequences with missing values.**

As both $\overline{T}_{i,m}$ and $\overline{T}_{j,m}$ have missing values, their means $(\mu_i, \mu_j)$ and standard deviations $(\sigma_i, \sigma_j)$ can be arbitrary values. We have:

$$d_{i,j} \geq \min_{\mu_j, \sigma_j} \sqrt{\min_{\mu_i, \sigma_i} \sum_{k \in R}\left(\frac{t_{i+k-1}-\mu_i}{\sigma_i} - \frac{t_{j+k-1}-\mu_j}{\sigma_j}\right)^2} = d_{i,j}^{LB} \quad (5)$$

Here $R = \{k \mid 1 \leq k \leq m, t_{i+k-1} \neq NaN, t_{j+k-1} \neq NaN\}$ is the intersection of the real-valued locations within $\overline{T}_{i,m}$ and $\overline{T}_{j,m}$ (see **Figure 7**). A diligent reader may have noticed that the lower bound expression $d_{i,j}^{LB}$ in (5) subsumes (1), the lower bound expression in Case 2. We can visualize this in **Figure 7**: if we remove $M_j$, the problem is transformed to Case 2. Therefore, we can directly substitute (3), the result of Case 2, into (5):

$$d_{i,j}^{LB} = \begin{cases} \sigma_j^R \sqrt{|R|} \min_{\sigma_j} \frac{1}{\sigma_j} & if\ q_{i,j} \leq 0 \\ \sigma_j^R \sqrt{|R|(1-q_{i,j}^2)} \min_{\sigma_j} \frac{1}{\sigma_j} & if\ q_{i,j} > 0 \end{cases} \quad (6)$$

We can see from (6) that $d_{i,j}^{LB}$ is dependent on $\sigma_j$ (controlled by the missing part of $\overline{T}_{j,m}$ in **Figure 7**): the larger $\sigma_j$, the smaller $d_{i,j}^{LB}$. Note that $\sigma_j$ can be as large as $+\infty$; in that case $d_{i,j}^{LB}$ becomes zero. This is a very undesirable lower bound, as *any* pair of missing value subsequences can be reported as a motif, even if they look very different from each other. **Figure 8** shows an example of this.

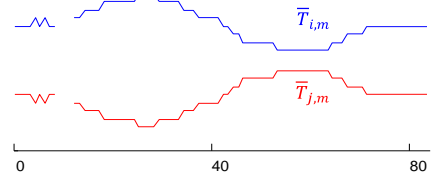

**Figure 8: Two subsequences with missing values. The real-valued parts of the subsequences look very different from each other, but if we fill the missing parts with infinitely large numbers, the z-normalized Euclidean distance of the two subsequences will become zero.**

Fortunately, sensor readings normally have physical limits. The accelerometer values on an iPhone 7 are limited to $\pm$ 8g ($\pm$ 78.48 m/s$^2$); virtually all medical sensors come with carefully specified limits to meet regulations (i.e., EU directive 93/42/EEC mandates that a pediatric lung ventilator monitor produces values in the range of 0 to 125$_{cmH2O}$), etc. Therefore, we can assume that the missing values in $\overline{T}_{j,m}$ are bounded by $[V_{min}, V_{max}]$. With this bound, we can derive the following inequality for $\sigma_j^2$ (we refer the interested reader to [18] for the complete derivation):

$$\sigma_j^2 \leq \frac{C^2}{4} + \frac{\sum_{k \in R_j} t_{j+k-1}^2 + |R_j|\left(B - \mu_j^{R_j}A\right)}{m} \quad (7)$$

Here $R_j$ is a set of the locations of all the real values within $\overline{T}_{j,m}$ (see **Figure 7**), $\mu_j^{R_j}$ is the mean of the real-valued part of $\overline{T}_{j,m}$, $C = V_{max} - V_{min}$, $B = V_{max}V_{min}$, $A = V_{max} + V_{min}$. In practice, we set $V_{min}$ and $V_{max}$ as the minimum and maximum value of the real-valued part of $\overline{T}_{j,m}$.

We can now evaluate $d_{i,j}^{LB}$ by substituting (7) back into (6):

$$d_{i,j}^{LB} = \begin{cases} \sqrt{|R|f_{LB}(j)} & if\ q_{i,j} \leq 0 \\ \sqrt{|R|(1-q_{i,j}^2)f_{LB}(j)} & if\ q_{i,j} > 0 \end{cases} \quad (8)$$

Here $f_{LB}(j) = \sigma_j^{R^2} / \left[\frac{C^2}{4} + \frac{\sum_{k \in R_j} t_{j+k-1}^2 + |R_j|\left(B - \mu_j^{R_j}A\right)}{m}\right]$. Note that $f_{LB}(j)$ is based on $R_j$, the real-valued part of $\overline{T}_{j,m}$ (recall **Figure 7**). However, as in Case 3 both $\overline{T}_{i,m}$ and $\overline{T}_{j,m}$ have missing values, we can analogously derive a similar lower bound expression as (8) based on $\overline{T}_{i,m}$, and set the larger expression as $d_{i,j}^{LB}$.

Formally, let $f_{LB}(p) = \sigma_p^{R^2} / \left[\frac{C^2}{4} + \frac{\sum_{k \in R_p} t_{p+k-1}^2 + |R_p|\left(B - \mu_p^{R_p}A\right)}{m}\right]$, then the lower bound distance for Case 3 is:

$$d_{i,j}^{LB} = \begin{cases} \sqrt{|R| \, max(f_{LB}(i), f_{LB}(j))} & if \ q_{i,j} \le 0 \\ \sqrt{|R| \, (1 - q_{i,j}^2) max(f_{LB}(i), f_{LB}(j))} & if \ q_{i,j} > 0 \end{cases} \quad (9)$$

The analysis of Case 3 is now complete. Finally, for completeness, let us briefly discuss Case 1.

### 3.3.3 Case 1

As neither $\overline{T}_{i,m}$ nor $\overline{T}_{j,m}$ contain any missing value in Case 1, we set $d_{i,j}^{LB}$ as the exact z-normalized Euclidean distance between $\overline{T}_{i,m}$ and $\overline{T}_{j,m}$, using the following equation ([26]):

$$d_{i,j}^{LB} = \sqrt{2m(1 - q_{i,j})} \quad (10)$$

Note that all three cases use the same expression of $q_{i,j}$ in (4).

Now that we have the lower bound distance for any subsequence pair in $\overline{T}$, we can also evaluate the lower bound matrix profile.

## 3.4 The MDMS Algorithm

The STOMP algorithm [26] can obtain the matrix profile of a time series that is free of missing values, in $O(n^2)$ time with only $O(n)$ space; as we will now show, in the face of missing data, our MDMS algorithm can obtain the lower-bound matrix profile with the same time and space complexity. TABLE I summarizes the algorithm.

TABLE I. MDMS ALGORITHM

| | |
|---|---|
| **Algorithm *MDMS(T, m)*** | |
| Input: A missing value time series *T*, subsequence length *m* | |
| Output: Lower bound matrix profile *P* and the associated lower bound matrix profile index *I of T* | |
| 1 | $n \leftarrow Length(T), len \leftarrow n-m+1$ |
| 2 | $vmax \leftarrow SlidingMax(T), vmin \leftarrow SlidingMin(T)$ |
| 3 | $Z \leftarrow PadZero(T), B \leftarrow OneZero(T), X \leftarrow ElementWiseSquare(Z)$ |
| 4 | $\mu z, \sigma z \leftarrow ComputeMeanStd(Z, m)$     // see [14] |
| 5 | $\mu b, \sigma b \leftarrow ComputeMeanStd(B, m)$     // see [14] |
| 6 | $QZ \leftarrow SlidingDotProduct(Z[1:m], Z), QZ\_first \leftarrow QZ$ //see [26] |
| 7 | $QB \leftarrow SlidingDotProduct(B[1:m], B), QB\_first \leftarrow QB$ //see [26] |
| 8 | $BZ \leftarrow SlidingDotProduct(B[1:m], Z), BZ\_first \leftarrow BZ$ //see [26] |
| 9 | $ZB \leftarrow SlidingDotProduct(Z[1:m], B), ZB\_first \leftarrow ZB$ //see [26] |
| 10 | $BX \leftarrow SlidingDotProduct(B[1:m], X), BX\_first \leftarrow BX$ //see [26] |
| 11 | $XB \leftarrow SlidingDotProduct(X[1:m], B), XB\_first \leftarrow XB$ //see [26] |
| 12 | $P \leftarrow \textbf{\textit{CalculateLBDistance}}\ (n, m, vmax, vmin, QZ, QB, BZ, ZB, BX, XB, \mu z, \sigma z, \mu b, \sigma b, i), I \leftarrow ones$     // initialization |
| 13 | **for** $i = 2$ **to** $len$     // in-order evaluation |
| 14 |    **for** $j = len$ **downto** $2$     // update dot product, see [26] |
| 15 |      $QZ[j] \leftarrow QZ[j-1]-Z[i-1]\times Z[j-1]+Z[i+m-1]\times Z[j+m-1]$ |
| 16 |      $QB[j] \leftarrow QB[j-1]-B[i-1]\times B[j-1]+B[i+m-1]\times B[j+m-1]$ |
| 17 |      $BZ[j] \leftarrow BZ[j-1]-B[i-1]\times Z[j-1]+B[i+m-1]\times Z[j+m-1]$ |
| 18 |      $ZB[j] \leftarrow ZB[j-1]-Z[i-1]\times B[j-1]+Z[i+m-1]\times B[j+m-1]$ |
| 19 |      $BX[j] \leftarrow BX[j-1]-B[i-1]\times X[j-1]+B[i+m-1]\times X[j+m-1]$ |
| 20 |      $XB[j] \leftarrow XB[j-1]-X[i-1]\times B[j-1]+X[i+m-1]\times B[j+m-1]$ |
| 21 |    **end for** |
| 22 |    $QZ[1] \leftarrow QZ\_first[i], QB[1] \leftarrow QB\_first[i], BZ[1] \leftarrow ZB\_first[i]$ |
| 23 |    $ZB[1] \leftarrow BZ\_first[i], BX[1] \leftarrow XB\_first[i], XB[1] \leftarrow BX\_first[i]$ |
| 24 |    $D \leftarrow \textbf{\textit{CalculateLBDistance}}\ (n, m, vmax, vmin, QZ, QB, BZ, ZB, BX, XB, \mu z, \sigma z, \mu b, \sigma b, i)$ |
| 25 |    $P, I \leftarrow ElementWiseMin(P, I, D, i)$ |
| 26 | **end for** |
| 27 | **return** $P, I$ |

Before discussing the algorithm in detail, we first need to introduce three important auxiliary time series (shown in line 3), $Z$, $X$ and $B$.

- We define $Z = z_1, z_2, \ldots z_n$, such that $z_i = t_i$ if $t_i \ne NaN$, and $z_i = 0$ if $t_i = NaN$. We can simply obtain $Z$ by filling zeros in the locations of $\overline{T}$ where the data is missing.

- We define $X = x_1, x_2, \ldots x_n = z_1^2, z_2^2, \ldots z_n^2$.

- We define $B = b_1, b_2, \ldots b_n$, such that $b_i = 1$ if $t_i \ne NaN$, and $b_i = 0$ if $t_i = NaN$. We can see that $B$ indicates the locations of the real-valued numbers and missing values in $\overline{T}$.

With these three auxiliary time series and the techniques introduced in [14][26], we can evaluate any lower bound distances introduced in the last section in $O(1)$ time with $O(n)$ space.

The MDMS algorithm is very similar to the STOMP framework [26]. In line 2 we evaluate the maximum and minimum values of the real-value part of every subsequence in *T*. Lines 4-5 evaluate the mean and standard deviation of every subsequence in *Z* and *B*, lines 6-11 initialize all the dot product vectors. We initialize the lower bound matrix profile *P* and matrix profile index *I* in line 12. Lines 13-26 iteratively evaluate the lower bound distance profile *D*, and update *P* and *I* if necessary. The *CalculateLBDistance* algorithm in lines 12 and 24 is shown in TABLE II.

TABLE II. ALGORITHM TO CALCULATE LOWERBOUND DISTANCE PROFILE

| | |
|---|---|
| **Algorithm *CalculateLBDistance* ($n, m, vmax, vmin, QZ, QB, BZ, ZB, BX, XB, \mu z, \sigma z, \mu b, \sigma b, i$)** | |
| Input: the length *n* of time series *T*, the subsequence length *m*, the maximum/minimum possible value vector *vmax/vmin*, dot product vectors *QZ, QB, BZ, ZB, BX, XB*, means and standard deviations $\mu z, \sigma z, \mu b, \sigma b$ of time series *Z* and *B*, current subsequence index *i* | |
| Output: Lower bound distance profile *D* | |
| 1 | **for** $j = 1$ **to** $n-m+1$ |
| 2 |    $ui \leftarrow ZB[j] / QB[j],\ uj \leftarrow BZ[j] / QB[j]$     $//\mu_i^R\ and\ \mu_j^R$ |
| 3 |    $vi \leftarrow XB[j] / QB[j] - ui^{\wedge}2,\ vj \leftarrow BX[j] / QB[j] - uj^{\wedge}2\ //\sigma_i^R\ and\ \sigma_j^R$ |
| 4 |    $q \leftarrow (QZ[j] / QB[j] - ui \times uj) / sqrt(vi \times vj)$     // (4) |
| 5 |    **if** $QB[j] == m$ **then**     // Case 1, $|R| = m$ |
| 6 |      $D[j] \leftarrow 2 \times m \times (1-q)$     // (10) |
| 7 |    **else** |
| 8 |      **if** $max(\mu b[i], \mu b[j]) == 1$ **then**     //Case 2 |
| 9 |        **if** $\mu b[i] > \mu b[j]$ **then** $vo \leftarrow vi, v \leftarrow \sigma z[i]^{\wedge}2$ |
| 10 |        **else** $vo \leftarrow vj, v \leftarrow \sigma z[j]^{\wedge}2$ |
| 11 |        **end if** |
| 12 |        **if** $q<=0$ **then** $D[j] \leftarrow QB[j] \times vo / v$     // (3) |
| 13 |        **else** $D[j] \leftarrow QB[j] \times vo / v \times (1 - q^{\wedge}2)$ |
| 14 |      **end if** |
| 15 |      **else**     //Case 3 |
| 16 |        $v1 \leftarrow vmax[i], v2 \leftarrow vmin[i], C \leftarrow v1-v2, B \leftarrow v1 \times v2, A \leftarrow v1+v2$ |
| 17 |        $ur \leftarrow \mu z[i]/\mu b[i], vr \leftarrow (\sigma z[i]^{\wedge}2 + \mu z[i]^{\wedge}2)/\mu b[i] - ur^{\wedge}2$ |
| 18 |        $f1 = vi/(ub[i] \times (vr+B+ur \times (ur-A)) + C^{\wedge}2/4)$ |
| 19 |        $v1 \leftarrow vmax[j], v2 \leftarrow vmin[j], C \leftarrow v1-v2, B \leftarrow v1 \times v2, A \leftarrow v1+v2$ |
| 20 |        $ur \leftarrow \mu z[j]/\mu b[j], vr \leftarrow (\sigma z[j]^{\wedge}2 + \mu z[j]^{\wedge}2)/\mu b[j] - ur^{\wedge}2$ |
| 21 |        $f2 = vj/(ub[j] \times (vr+B+ur \times (ur-A)) + C^{\wedge}2/4)$ |
| 22 |        **if** $q<=0$ **then** $D[j] \leftarrow QB[j] \times max(f1, f2)$     // (9) |
| 23 |        **else** $D[j] \leftarrow QB[j] \times max(f1, f2) \times (1 - q^{\wedge}2)$ |
| 24 |      **end if** |
| 25 |      **end if** |
| 26 |    **end if** |
| 27 | **end for** |
| 28 | **return** *D* |

The *CalculateLBDistance* algorithm evaluates all $n-m+1$ lower bound distance values in *D* with equations (3), (9) and (10). Line 4 evaluates (4). Case 1 is handled in lines 5-6. Lines 8-14 handle Case 2, and Case 3 is evaluated in lines 15-24.

We can see that each loop of the *CalculateLBDistance* algorithm in TABLE II (lines 2-26) can be evaluated in $O(1)$ time, so the time complexity of *CalculateLBDistance* is $O(n)$. The space needed to store all the vectors in TABLE I is $O(n)$, and each loop in lines 14-25 of TABLE I takes $O(n)$ time. Therefore, the time complexity of MDMS algorithm is $O(n^2)$ and the space complexity is $O(n)$, the same as STOMP [26].

Furthermore, we can see that unlike most imputation algorithms [21], our MDMS algorithm is extremely model-free and parameter-free. The only inputs to the algorithm are the time series and the subsequence length. In the next section, we will use two case studies to demonstrate the robustness and efficacy of our ultra-fast, parameter-free motif discovery algorithm in the face of missing data.

## 4 EXPERIMENTAL EVALUATION

We begin by noting that all the code and data used in this work are archived in perpetuity at [18]. In the following two case studies, we consider both random-missing data (data with sparsely located, random missing timestamps) and block-missing data (data with consecutive missing timestamps).

For each case study, we compare our method with the commonly used strawman in the literature to handle missing data: linear imputation, as shown in **Figure 9**.

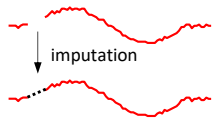

**Figure 9: To evaluate our method, we compare our result with that of linear imputation.**

### 4.1 Case Study: Seismological Data

Repeated pattern (i.e. *motif*) discovery is a fundamental tool in seismology, which allows for the discovery of foreshocks, triggered earthquakes, swarms, volcanic activity, and induced seismicity [25]. However, this domain is replete with missing data.

For example, a classic paper notes "*A frequent dilemma in spectral analysis* (in seismology) *is the incompleteness of the data record, either in the form of occasional missing data or as larger gaps*" (our emphasis) [1]. In this experiment, we demonstrate that we can handle both cases.

We consider a dataset for which we know the answer from external sources. On April 30th, 1996, there was an earthquake of magnitude 2.12 in Sonoma County, California. Then, on December 29th, 2009, about 13.6 years later, there was another earthquake with a similar magnitude. To allow the results to be visualized in a single plot, we edited this data such that the two earthquakes happen just 15 seconds apart. We set the subsequence length as 2,000, which corresponds to one second of data. As shown in **Figure 10** (red curve), when there is no missing data the matrix profile correctly discovers the locations of the two earthquakes.

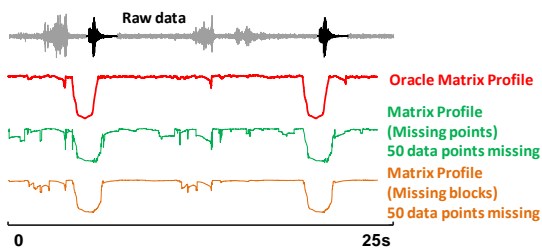

**Figure 10: A raw seismograph contrived such that two earthquakes from the same region happen 15 seconds apart. The matrix profile computed with no missing data (red curve) finds the true event, as does MDMS even in the presence of missing points (green curve) or missing blocks (orange curve).**

To test our algorithm for the "occasional missing data" case, we randomly deleted 50 data points. As **Figure 10** (green curve) shows, the matrix profile is still minimized at the correct location, and there are no false positives (no other small values in the matrix profile besides the two deep valleys). This shows the robustness of our algorithm in the face of random missing data.

Next, we consider the "larger gaps" (or the block-missing data) case. Here, instead of removing individual data points, we removed two consecutive blocks of length 25. As shown in **Figure 10** (orange curve), the shape of the lower bound matrix profile still looks very similar to that of the oracle matrix profile (red curve). We see only two deep valleys at the vicinity of the motifs, so no false positive patterns are discovered.

The result demonstrates that our lower bound matrix profile is robust against producing false positives.

To test the robustness of our algorithm against false negatives, we removed a block of missing data at the center of the second earthquake pattern. The length of the missing block is 400, which is 20% of the subsequence length. In **Figure 11**, we compared our lower-bound matrix profile result with the matrix profile generated by linear imputation.

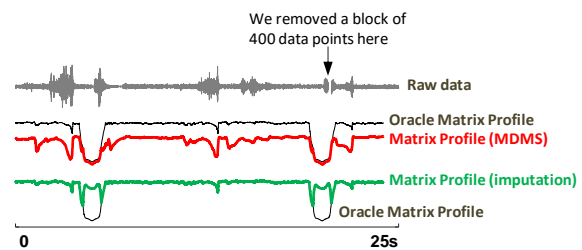

**Figure 11: We removed 400 consecutive data points at the center of the second earthquake pattern. The oracle matrix profile computed with no missing data (black curve) finds the true event, as does MDMS (red curve) even in the presence of a large missing block. The Matrix Profile generated after linear imputation (green curve) fails to capture the minimum points within the oracle matrix profile.**

We can see that the lower-bound matrix profile generated by our MDMS algorithm (red curve) agrees closely with the oracle matrix profile (black curve) at the vicinity of the two earthquake patterns, while the matrix profile generated after linear imputation (green curve) shows a high value at these locations.

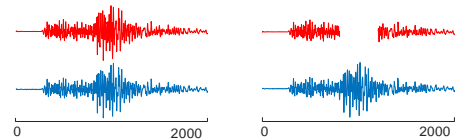

**Figure 12: The first motif found by the MDMS algorithm (right) in the presences of a large missing block is identical with the first motif found in the oracle data (left).**

As a result, the MDMS algorithm successfully captures the 1st motif (as shown in **Figure 12**) even in the presence of a large missing block within one earthquake pattern, while the imputation method misses the 1st motif within the oracle data. This illustrates two major strength of our algorithm over imputation methods. Firstly, our algorithm does not allow false negatives. Secondly, our algorithm is more robust to large missing blocks as it does not change the data, while imputation method can change the data a lot. In the next case study, we will further demonstrate the robustness of MDMS in the presence of missing blocks.

### 4.2 Case Study: Activity Data from Video

Time series extracted from video often has missing data reflecting "frame drops" due to bandwidth congestion [6]. To test our algorithm in this context, we examine the activity dataset of [20]. This dataset consists of a 13.3 minute 10-fps video sequence of an actor performing various activities. From this data, the original authors extracted 721 channels of the optical flow time series, and the length of each time series is 8,000. We consider the time series corresponding to the 533[rd] channel, which is suggestive of the structure in places but is noisy. The data is shown in **Figure 13**.*top*, the subsequence length is 120. From the oracle matrix profile in **Figure 13** (black curve), we can extract the 1[st] motif in the oracle data. To test the performance of the MDMS algorithm, here we remove 12 consecutive data points in the center of one of the 1[st] motif patterns. In **Figure 13**, we compare our lower-bound matrix profile result (red curve) with the matrix profile generated by linear imputation (green curve).

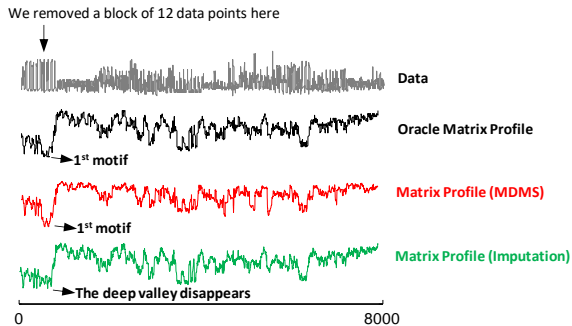

**Figure 13: A raw activity time series. We removed 12 consecutive data points in one of the 1[st] motif patterns in the time series. The oracle matrix profile computed with no missing data (black curve) finds the true motif starting at the 540[th] and the 622[nd] data points. With the presence of 12 missing data points, the MDMS algorithm finds the same motif as the oracle result (red curve), starting at 520[th] and 602[nd] data points. The Matrix Profile generated after linear imputation (green curve) fails to capture the two deep valleys within the oracle matrix profile and thus misses the 1[st] motif.**

We can see that the oracle matrix profile (black curve) shows two apparent valleys at the locations of the 1[st] motif, as does the lower bound matrix profile generated by MDMS (red curve). The 1[st] motif discovered by the MDMS algorithm (shown in **Figure 14**.*right*) is identical to the oracle motif (shown in **Figure 14**.*left*). In contrast, the matrix profile generated by imputation does not have these valleys, and thus misses the 1[st] motif of the oracle data.

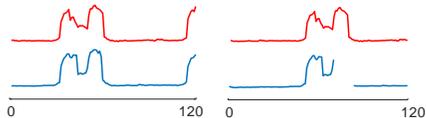

**Figure 14: The first motif found by the MDMS algorithm (*right*) is identical to the first motif within the oracle data (*left*), despite a small phase shift.**

We can see from **Figure 14** that though a large portion of the blue pattern is missing, our MDMS algorithm still finds it very similar with the red pattern. This example further demonstrates that our algorithm is robust against missing true motif patterns.

## 4.3 Quantifying the Robustness of MDMS

As MDMS evaluates the lower bound matrix profile, it naturally does not allow false negatives, but it can produce false positives. Here we perform two "stress tests" to evaluate the robustness and limitations of our MDMS algorithm against false positives.

We use the seismograph data in **Figure 10** again for the stress test. The subsequence length in this dataset is $m$=2,000.

We first test the sensitivity of MDMS over the length of missing blocks. Here we remove two missing blocks of length $p$, located at 7.5s and 15s respectively, from the data. In **Figure 15**, we show how the lower bound matrix profile varies as we increase $p$.

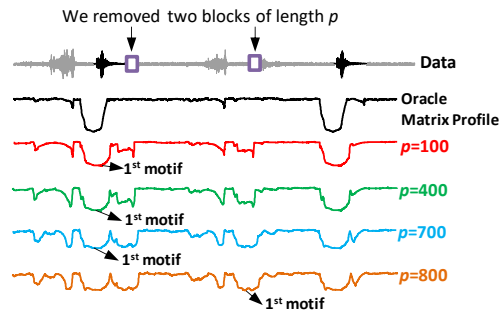

**Figure 15: Lower bound matrix profiles corresponding to different missing block lengths. We removed 2 blocks of length $p$ from the seismograph. The oracle matrix profile (black curve) finds the true motif. For $p$=100, $p$=400 and $p$=700, MDMS is able to find the true event as the 1[st] motif. When $p$=800, MDMS finds a false positive as the 1[st] motif.**

Note that the removed blocks are *not* within the two repeated earthquake patterns. As a result, the lower bound matrix profiles are the same as the oracle matrix profile at the vicinity of the two repeated patterns, while lower at other locations. In other words, it is easier to detect false positives with such setting.

We can see that when $p$=100 (5% of the subsequence length $m$), the side valleys in the oracle Matrix Profile become deeper, and two more side valleys show up at the vicinity of the removed blocks. As $p$ increases, all the side valleys become deeper and deeper. For $p$=100, $p$=400 and $p$=700, we are able to find the true event as the 1[st] motif with MDMS. However, when $p$=800 (40% of $m$), the 1[st] motif (corresponding to the minimum point of the lower bound matrix profile) is no longer the true event. We show this false positive motif pair in **Figure 16**.

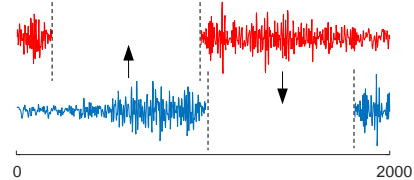

**Figure 16: The 1[st] motif found by the MDMS algorithm when $p$=800.**

The two subsequences are both at the vicinity of the second missing block in **Figure 15**. With a close inspection we can see why this pair is reported as the 1[st] motif by MDMS. If we fill in the missing part of each subsequence with their counterpart in the other subsequence (shown by the arrows in **Figure 16**), the two can be *very* similar to each other. Since MDMS does not allow false negatives, it will capture and report this possible matching pair.

Furthermore, **Figure 16** implies that $p$ cannot be larger than 50% of the subsequence length, otherwise MDMS will be able to find a perfect match (with one subsequence missing the first 50% and the other missing the second 50%). When $p > 700$ (35% of the subsequence length), we are already very close to this limit. Therefore, in **Figure 15** we can see very deep side valleys at the vicinity of the missing blocks, and we are prone to detect false positives. When $p \leq 400$ (20% of the subsequence length), the two main valleys corresponding to the true events dominate, so we will not detect false positives.

We have demonstrated that MDMS is robust against discovering false positives when there are two missing blocks, and when the length of the two blocks are within a reasonable range. Next, we "stress test" the sensitivity of MDMS over the total number of missing values in the data. We again use the seismograph dataset in **Figure 10**, which consist of 50,000 data points. The lower bound matrix profile results are shown in **Figure 17**.

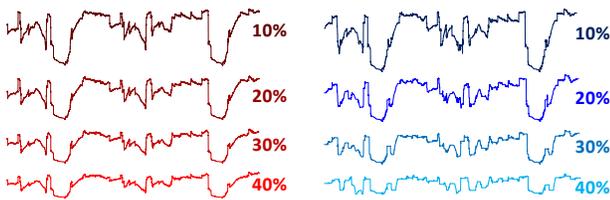

**Figure 17: Lower bound matrix profiles corresponding to various percentage of data missing.** *left)* random-missing data *right)* block-missing data.

In the first run, we randomly selected 5,000, 10,000, 15,000 and 20,000 points to remove from the data. From **Figure 17**.*left*, we can see that the scale of the matrix profile decreased as more points are missing. However, even when 40% of the data is missing, the two valleys corresponding to the true events (recall **Figure 10**) still dominate. In the second run, we removed blocks of length 200 from the data. The missing blocks were uniformly distributed, and the number of missing blocks increased from 25 to 100. **Figure 17**.*right* shows that even when 30% of the data is missing, the two main valleys still dominate. When 40% of the data is missing, the two main valleys are no longer apparent, but we can still find the true events as the 1$^{st}$ motif. The experiment demonstrates that MDMS is robust against detecting false positives even if a large percentage of data is missing.

## 5 CONCLUSIONS

We introduced what we believe to be the first time series motif discovery algorithm that can find motifs in the presence of missing data. The algorithm has the same time and space complexity as the fastest known algorithm for motif discovery [26]. We formally proved the admissibility of our algorithm, it does not produce any false negatives. Experimental results show that our algorithm is also robust against false positives even when a large portion of the data is missing. Because our algorithm is based on creating a special version of the matrix profile [22][26], our work may have implications for other algorithms that can exploit the matrix profile, including discord discovery and time series joins. The lower bounds introduced can also be used to accelerate various length motif discovery. We leave such considerations to future work.